\documentclass[sigconf]{acmart}

\usepackage{tcolorbox}
\usepackage{listings}
\usepackage[table]{xcolor}
\usepackage{afterpage}
\usepackage{booktabs}
\usepackage{tabularx}
\usepackage{array}
\usepackage{enumitem}
\usepackage{hyperref}
\newcommand{\first}[1]{\cellcolor{blue!30}#1}
\newcommand{\second}[1]{\cellcolor{blue!20}#1}
\newcommand{\third}[1]{\cellcolor{blue!10}#1}
\newcommand{\redfirst}[1]{\cellcolor{red!30}#1}
\newcommand{\redsecond}[1]{\cellcolor{red!20}#1}
\newcommand{\redthird}[1]{\cellcolor{red!10}#1}
\newcommand{\bfirst}[1]{\colorbox{blue!30}#1}
\newcommand{\bsecond}[1]{\colorbox{blue!20}#1}
\newcommand{\bthird}[1]{\colorbox{blue!10}#1}
\newcommand{\rfirst}[1]{\colorbox{red!30}#1}

\newcommand{\CFV}{\textsf{CFVBench}}

\tcbuselibrary{skins,breakable,listings}

\newtcblisting{markdownbox}{
    breakable,
    enhanced,
    colback=lightgray,
    colframe=black,
    boxrule=1pt,
    arc=3pt,
    left=8pt,
    right=8pt,
    top=8pt,
    bottom=8pt,
    fonttitle=\bfseries,
    before skip=10pt,
    after skip=10pt,
    listing only,
    listing options={
        basicstyle=\ttfamily\footnotesize,
        breaklines=true,
        breakatwhitespace=true,
        tabsize=2,
        showstringspaces=false,
        columns=flexible,
        xleftmargin=0pt,
        xrightmargin=0pt,
        frame=none,
        aboveskip=0pt,
        belowskip=0pt,
        lineskip=0pt
    }
}

\definecolor{lightgray}{RGB}{245,245,245}
\definecolor{softblue}{RGB}{70,130,180}
\definecolor{softgreen}{RGB}{60,150,90}
\definecolor{softred}{RGB}{180,80,80}
\definecolor{softpurple}{RGB}{140,90,160}
\definecolor{softorange}{RGB}{200,120,60}
\definecolor{lightblue}{RGB}{235,245,255}
\definecolor{lightgreen}{RGB}{240,250,245}
\definecolor{lightred}{RGB}{255,245,245}
\definecolor{highlight}{HTML}{e4f0ee}
\colorlet{secondcolor}{blue!20}

\usepackage{graphicx} 
\usepackage{tikz} 
\usepackage{xcolor}  

\newcommand{\greencheck}{\textcolor{green!70!black}{\checkmark}}
\newcommand{\redcross}{\textcolor{red!70!black}{$\times$}}

\DeclareRobustCommand{\ice}{
  \IfFileExists{pictures/ice.png}{
    \raisebox{-0.5ex}{\includegraphics[height=2.0ex]{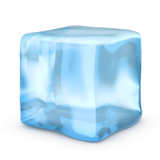}}%
  }{
    \raisebox{-0.5ex}{\begin{tikzpicture}[x=1ex,y=1ex,scale=0.17,baseline=-0.5ex]\draw[softgreen,fill=white] (0,3) -- (2,5) -- (4,3) -- (2,1) -- cycle; \draw[softgreen] (2,0) -- (2,6); \draw[softgreen] (0,3) -- (4,3);\end{tikzpicture}}%
  }%
}

\DeclareRobustCommand{\swaparrow}{%
  \IfFileExists{pictures/swaparrow.png}{%
    \raisebox{-0.5ex}{\includegraphics[height=2.0ex]{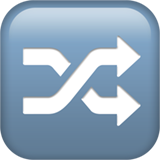}}%
  }{
    \raisebox{-0.5ex}{\begin{tikzpicture}[x=1ex,y=1ex,scale=0.14,baseline=-0.5ex]\draw[->,line width=0.9pt,softgreen] (0,2) to[out=0,in=180] (3,2); \draw[<-,line width=0.9pt,softgreen] (0,0) to[out=0,in=180] (3,0);\end{tikzpicture}}%
  }%
}

\AtBeginDocument{%
  }

\setcopyright{acmlicensed}
\copyrightyear{2018}
\acmYear{2018}
\acmDOI{XXXXXXX.XXXXXXX}
\acmConference[Conference acronym 'XX]{Make sure to enter the correct
  conference title from your rights confirmation email}{June 03--05,
  2018}{Woodstock, NY}
\acmISBN{978-1-4503-XXXX-X/2018/06}

\begin{document}

\title{CFVBench: A Comprehensive Video Benchmark for Fine-grained Multimodal Retrieval-Augmented Generation}


\author{Kaiwen Wei\textsuperscript{*}}
\email{weikaiwen@cqu.edu.cn}
\affiliation{
  \institution{Chongqing University}
  \city{Chongqing}
  \country{China}
}

\author{Xiao Liu\textsuperscript{*}}
\email{Liu-xiao-@outlook.com}
\affiliation{
  \institution{Chongqing University}
  \city{Chongqing}
  \country{China}
}

\author{Jie Zhang\textsuperscript{*}}
\email{zhang15827347943@163.com}
\affiliation{
  \institution{Independent Researcher}
  \country{China}
}

\author{Zijian Wang\textsuperscript{*}}
\email{wangzijian@stu.cqu.edu.cn}
\affiliation{
  \institution{Chongqing University}
  \city{Chongqing}
  \country{China}
}

\author{Ruida Liu}
\email{dang9ab3e5_1@163.com}
\affiliation{
  \institution{Chongqing University}
  \city{Chongqing}
  \country{China}
}

\author{Yuming Yang}
\email{ymyang@cqu.edu.cn}
\affiliation{
  \institution{Chongqing University}
  \city{Chongqing}
  \country{China}
}

\author{Xin Xiao}
\email{20241401023@stu.cqu.edu.cn}
\affiliation{
  \institution{Chongqing University}
  \city{Chongqing}
  \country{China}
}

\author{Xiao Sun}
\email{sunx@stu.cqu.edu.cn}
\affiliation{
  \institution{Chongqing University}
  \city{Chongqing}
  \country{China}
}

\author{Haoyang Zeng}
\email{202514156683@stu.cqu.edu.cn}
\affiliation{
  \institution{Chongqing University}
  \city{Chongqing}
  \country{China}
}

\author{Changzai Pan}
\email{panpanaqm@126.com}
\affiliation{
  \institution{Independent Researcher}
  \country{China}
}

\author{Yidan Zhang}
\email{zhangyidan19@mails.ucas.ac.cn}
\affiliation{
  \institution{University of the Chinese Academy of Sciences}
  \country{China}
}

\author{Jiang Zhong\textsuperscript{†}}
\email{zjstud@cqu.edu.cn}
\affiliation{
  \institution{Chongqing University}
  \city{Chongqing}
  \country{China}
}

\author{Peijin Wang\textsuperscript{†}}
\email{wangpeijin17@mails.ucas.ac.cn}
\affiliation{
  \institution{Aerospace Information Research Institute, Chinese Academy of Sciences}
  \country{China}
}

\author{Yingchao Feng\textsuperscript{†}}
\thanks{\textsuperscript{*}These authors contributed equally to this work.}
\thanks{\textsuperscript{†}Corresponding author.}
\email{fengyc@aircas.ac.cn}
\affiliation{
  \institution{Aerospace Information Research Institute, Chinese Academy of Sciences}
  \country{China}
}

\renewcommand{\shortauthors}{Trovato et al.}

\begin{abstract}
Multimodal Retrieval-Augmented Generation (MRAG) enables Multimodal Large Language Models (MLLMs) to generate responses with external multimodal evidence, and numerous video-based MRAG benchmarks have been proposed to evaluate model capabilities across retrieval and generation stages. However, existing benchmarks remain limited in modality coverage and format diversity, often focusing on single- or limited-modality tasks, or coarse-grained scene understanding. To address these gaps, we introduce \CFV, a large-scale, manually verified benchmark constructed from 599 publicly available videos, yielding 5,360 open-ended QA pairs. \CFV\ spans high-density formats and domains such as chart-heavy reports, news broadcasts, and software tutorials, requiring models to retrieve and reason over long temporal video spans while maintaining fine-grained multimodal information. Using \CFV, we systematically evaluate 7 retrieval methods and 14 widely-used MLLMs, revealing a critical bottleneck: current models (even GPT5 or Gemini) struggle to capture transient yet essential fine-grained multimodal details. To mitigate this, we propose Adaptive Visual Refinement (AVR), a simple yet effective framework that adaptively increases frame sampling density and selectively invokes external tools when necessary. Experiments show that AVR consistently enhances fine-grained multimodal comprehension and improves performance across all evaluated MLLMs.
\end{abstract}

\ccsdesc[500]{Information systems~Retrieval Augmented Generation}


\keywords{Multimodal Retrieval Augmented Generation, Video Understanding, Benchmark Evaluation, Multimodal Large Language Models
}

\received{20 February 2007}
\received[revised]{12 March 2009}
\received[accepted]{5 June 2009}

\maketitle

\section{Introduction}

\begin{figure}
    \centering
    \includegraphics[width=1\linewidth]{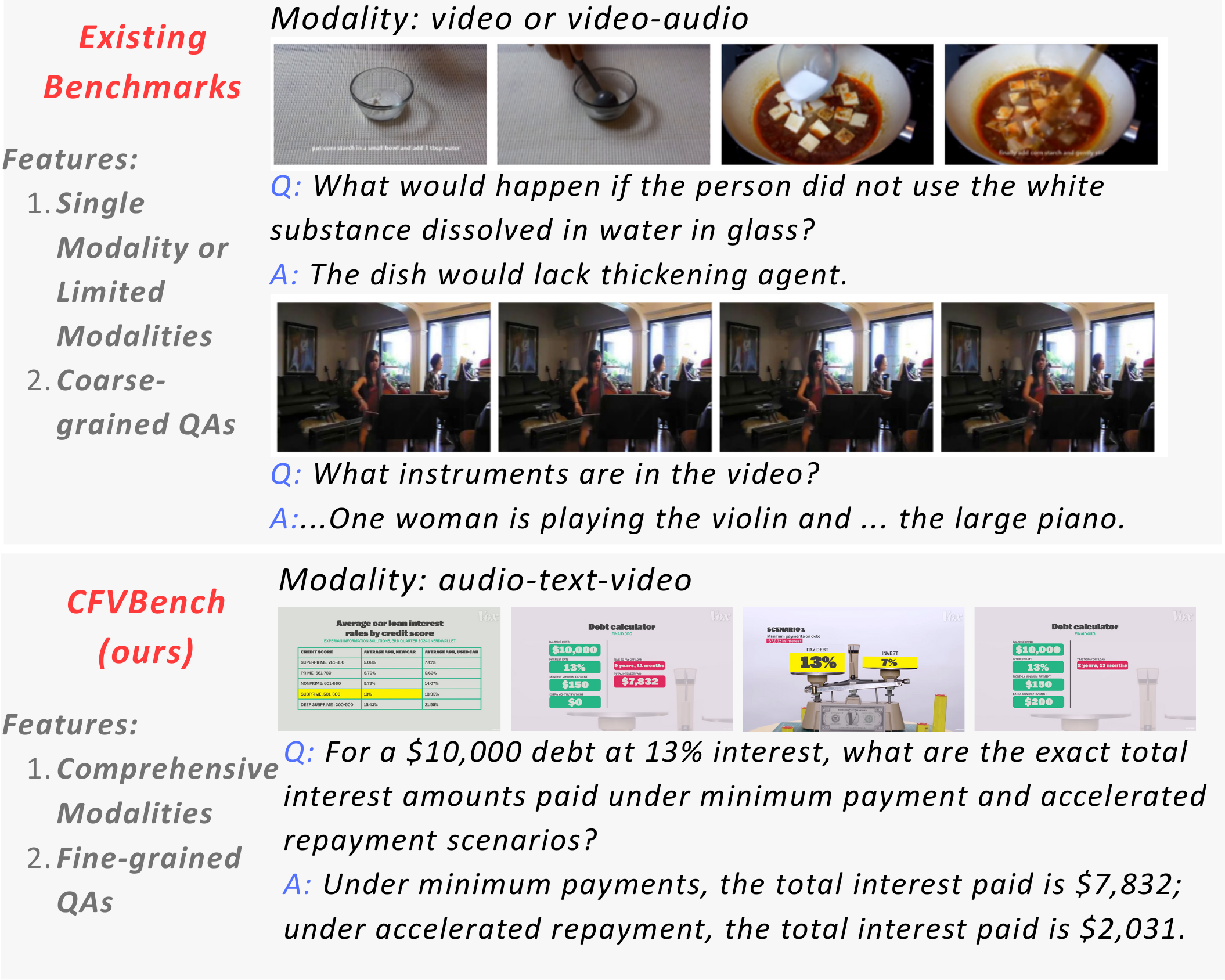}
    \caption{Comparison of video-based MRAG benchmarks with \CFV, where clues are embedded in tables/on-screen text of video frames, requiring fine-grained reasoning.}
    \label{fig:intro}
\end{figure}

In recent years, the Retrieval-Augmented Generation (RAG) paradigm has become indispensable for enhancing the factual accuracy, knowledge grounding~\cite{li2024enhancing}, and explainability~\cite{zhao2024retrieval} of Large Language Models (LLMs). This paradigm has been further extended to the Multimodal RAG (MRAG) framework, which leverages external multimodal knowledge (such as images, audio, and video) to address increasingly complex, real-world information-seeking tasks~\cite{mei2025survey,zhao2023retrieving}. Among these modalities, video-based MRAG~\cite{ma2021prop,zhang2021adversarial,sun2023learning,qu2024tiger,jiang2025kind} poses unique challenges due to its temporal dynamics, the tight coupling of visual and auditory streams, and the presence of dense, transient details~\cite{mao2025multi}.

\begin{table}[t]
\caption{Comparison of video-based MRAG benchmarks. MR: Multimodal Fine-grained Reasoning; RQA: Retrieval-based Question-Answering; IM: Intra-Modal Multi-hop Questions; OE: Open-Ended Questions; CMF: Comprehensive Modalities and Format (e.g., video, text, audio, chart, table, etc.). Notation / indicates that benchmark is not presented in QA format.}
\label{tab:dataset_comparison}
\renewcommand{\arraystretch}{0.9}
\resizebox{\columnwidth}{!}{%
\small
\begin{tabularx}{\columnwidth}{>{\centering\arraybackslash}p{0.35\columnwidth}*{6}{>{\centering\arraybackslash}p{0.08\columnwidth}}}
\toprule
\textbf{Dataset} & \textbf{MR} & \textbf{RQA} & \textbf{IM} & \textbf{OE} & \textbf{CMF} \\
\midrule
\multicolumn{6}{@{}l}{\textbf{\textit{Video Benchmarks}}} \\
\midrule
ShareGPT4Video \cite{chen2024sharegpt4video}& \redcross & \redcross & \redcross & / & \redcross \\
NExT-QA \cite{xiao2021next}& \redcross & \redcross & \redcross & \greencheck & \redcross \\
Video-MME \cite{fu2025video}& \redcross & \redcross & \greencheck & \redcross & \redcross \\
LongVideoBench \cite{wu2024longvideobench}& \redcross & \redcross & \greencheck & \redcross & \redcross \\
MovieChat \cite{song2024moviechat}& \redcross & \redcross & \greencheck & \greencheck & \redcross \\
EgoSchema \cite{mangalam2023egoschema}& \redcross & \redcross & \greencheck & \redcross & \redcross \\
MVBench \cite{li2024mvbench}& \redcross & \redcross & \redcross & \greencheck & \redcross \\
MMBench-Video \cite{fang2024mmbench}& \redcross & \redcross & \redcross & \greencheck & \redcross \\
SOK-Bench \cite{wang2024sok}& \redcross & \redcross & \redcross & \redcross & \redcross \\
\midrule
\multicolumn{6}{@{}l}{\textbf{\textit{Multimodal Benchmarks}}} \\
\midrule
UnAV-100 \cite{geng2023dense}& \redcross & \redcross & / & / & \redcross \\
VAST-27M \cite{chen2023vast}& \redcross & \redcross & \greencheck & / & \redcross \\
OmniBench \cite{li2024omnibench}& \greencheck & \redcross & \redcross & \redcross & \redcross \\
Cinepile \cite{rawal2024cinepile}& \redcross & \redcross & \redcross & \redcross & \redcross \\
Video-Bench \cite{ning2023video}& \redcross & \redcross & \redcross & \greencheck & \redcross \\
AVQA \cite{yang2022avqa}& \redcross & \redcross & \redcross & \redcross & \redcross \\
AVInstruct \cite{ye2024cat}& \redcross & \redcross & \redcross & \greencheck & \redcross \\
Music-AVQA2.0 \cite{liu2024tackling}& \redcross & \redcross & \redcross & \redcross & \redcross \\
VGGSound \cite{chen2020vggsound}& \redcross & \redcross & / & / & \redcross \\
AVHaystacks \cite{chowdhury2025magnet}& \redcross & \greencheck & \greencheck & \greencheck & \redcross \\
AVHBench \cite{sung2024avhbench}& \redcross & \redcross & \redcross & \redcross & \redcross \\
\rowcolor{secondcolor}
\textbf{\CFV\ (ours)} & \greencheck & \greencheck & \greencheck & \greencheck & \greencheck \\
\bottomrule
\end{tabularx}%
}
\end{table}

To evaluate the increasingly sophisticated capabilities of video-based MRAG systems, numerous benchmarks ~\cite{chen2023can,ye2024cat,li2024omnibench,chen2024sharegpt4video} have been proposed, typically focusing on tasks such as basic factual retrieval in visual questions, general video event understanding, or cross-modal reasoning. However, as illustrated in Fig.~\ref{fig:intro} (see full-size case in Appendix~\ref{appendix:full_size}) and Table~\ref{tab:dataset_comparison}, existing benchmarks face two key limitations when confronted with real-world complexity. (1) \textbf{limited modality and format coverage}: current benchmarks often involve restricted modality pairings (e.g., text-video or audio-video), overlooking datasets that simultaneously encompass audio-text-video streams and densely embedded domain-specific formats, such as data charts, complex tables, or dynamic on-screen text, which are critical for real-life video comprehension. (2) \textbf{insufficient fine-grained reasoning}: most benchmarks rely on coarse-grained analysis, such as scene classification or major entity recognition, and do not adequately evaluate a model’s ability to perform precise, detailed multimodal reasoning. For example,  extracting exact numerical values from rapidly changing charts or following sequential, understanding the complex operations and timing relationships in a tutorial, both of which are essential for grounded, factual multimodal question answering task.

To address these limitations, we introduce the Comprehensive Fine-grained Video-based retrieval-augmented generation Benchmark (\CFV), a large-scale, manually verified, multimodal dataset specifically designed to evaluate video-based MRAG systems. \CFV\  is constructed from real-world videos on YouTube. It contains 599 videos and 5,360 open-ended question-answering (QA) pairs, spanning high-density content such as chart-heavy reports, news broadcasts, and software tutorials. The benchmark requires models to retrieve information across long temporal video spans and multiple modalities while maintaining a high level of fine-grained visual fidelity. Using \CFV, we conduct a comprehensive evaluation of 7 popular retrieval methods and 14 publicly available Multimodal Large Language Models (MLLMs), revealing a critical bottleneck: \textit{existing methods (even GPT5 or Gemini) struggle significantly with fine-grained multimodal comprehension}. Further human evaluation shows that models frequently miss fine-grained visual cues and transient events, and exhibit internal issues such as hallucination, resulting in incomplete or inaccurate understanding.

%

Motivated by this critical finding, we propose the Adaptive Visual Refinement (AVR) framework. AVR is a simple yet effective, two-stage framework that enables MLLMs to assess their current information state and execute a resource-efficient refinement strategy. In the first stage, adaptive frame interpolation mechanism employs a scoring system to evaluate information sufficiency and automatically increase frame sampling density only when an incomplete evidence chain is detected. In the second stage, on-demand tool invocation strategy intelligently activates specialized external tools, such as Optical Character Recognition (OCR)~\cite{vedhaviyassh2022comparative} and object detectors~\cite{cheng2024yolo}, when necessary to extract embedded visual text or entity relevant to the query. Experimental results show that AVR effectively enhances the fine-grained multimodal comprehension of existing MLLMs, enabling them to capture subtle visual details and achieve state-of-the-art (SOTA) performance on the proposed \CFV. In summary, the main contributions of this work are:

1) We introduce \CFV, a large-scale MRAG benchmark (599 videos, 5,360 QA pairs) with high-density content (e.g., charts, tutorials), specifically designed to assess complex fine-grained multimodal comprehension across long temporal video spans.

2) We evaluate 7 retrieval methods and 14 MLLMs on \CFV, uncovering a key bottleneck of existing MLLMs in fine-grained multimodal comprehension, likely caused by internal hallucinations and insufficient attention to transient details.

3) We propose AVR, a simple yet effective framework that leverages adaptive frame interpolation and on-demand tool invocation to enhance fine-grained visual understanding and improve performance across all evaluated MLLMs. The dataset, code, and prompts will be released upon acceptance of the paper.

\section{Related Works}

\subsection{Multimodal RAG Benchmarks}
Existing Multimodal Retrieval-Augmented Generation (MRAG) benchmarks have progressively expanded the integration of diverse modalities~\cite{mei2025survey,zhao2023retrieving}, including text, images, audio, and video. Early knowledge-based visual question answering datasets, such as KB-VQA~\cite{wang2015explicit} and KVQA~\cite{shah2019kvqa}, rely on closed-domain knowledge, while OK-VQA~\cite{marino2019ok} and A-OKVQA~\cite{schwenk2022okvqa} incorporate external knowledge but mainly focus on single-step retrieval rather than complex reasoning. More recent benchmarks have broadened modality coverage and task complexity. MRAG-Bench~\cite{hu2024mrag} and MRAMG-Bench~\cite{yu2025mramg} combine text and images, Chart-MRAG Bench~\cite{yang2025benchmarking} integrates textual and chart data, ManyModalQA considers text, images, and tables, WavCaps~\cite{mei2024wavcaps}, and MusicCaps~\cite{agostinelli2023musiclm} emphasize audio-language understanding, KnowIT VQA~\cite{garcia2020knowit} and SOK-Bench~\cite{wang2024sok} focus on video reasoning, and InfoSeek~\cite{chen2023can} and Encyclopedic VQA~\cite{mensink2023encyclopedic} address knowledge-intensive, information-seeking tasks. Benchmarks with broader modality coverage, such as MultiModalQA~\cite{talmor2021multimodalqa} and AVHaystacks~\cite{chowdhury2025magnet}, still tend to simplify question formats through templates or multiple-choice settings (further comparison of the benchmarks are illustrated in Table \ref{tab:dataset_comparison}.) Despite these advancements, most existing benchmarks remain limited in modality and format diversity and lack systematic evaluation of fine-grained reasoning. To address this gap, we introduce CFVBench, a benchmark designed to evaluate fine-grained multimodal comprehension ability of different video-based MRAG systems.

\begin{figure*}
    \centering
    \includegraphics[width=0.9\linewidth]{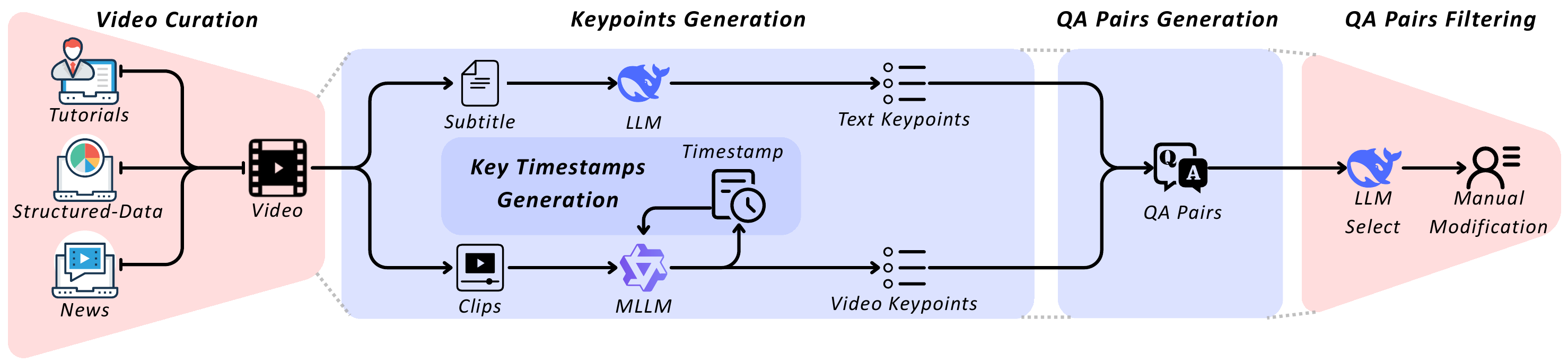}
    \caption{The dataset construction process of CFVBench.}
    \label{fig:pipeline}
\end{figure*}

\subsection{Multimodal RAG Methods}
Early MRAG methods typically converted multimodal data into unified textual representations for a text-based RAG pipeline~\cite{ma2021prop,zhang2021adversarial}, causing substantial information loss, while later approaches preserved original multimodal data and introduced MLLMs for cross-modal retrieval and generation~\cite{sun2023learning,zeng2024scalable}. Recent MRAG systems incorporate advanced modules such as multimodal search planners~\cite{lou2022audio,wang2024text}, interleaved text-image output modules~\cite{zhang2023irgen,li2025revolutionizing}, LLM-based cross-modal rerankers~\cite{qu2024tiger,lin2024draw}, and cross-modal refiners~\cite{wei2024promptmm,jiang2025kind} to achieve end-to-end multimodal interaction. Despite these advances, evaluation still relies heavily on VQA~\cite{li2024benchmarking,lin2023fine} or VidQA datasets~\cite{yu2019activitynet,sanabria2018how2}, limiting assessment of fine-grained reasoning. By conducting experiments on the proposed CFVBench, we find that existing video-based MRAG methods struggle with detailed multimodal comprehension, which motivates the Adaptive Visual Refinement (AVR) framework that evaluates retrieved visual information and selectively enhances frame sampling or invokes specialized tools to improve fine-grained comprehension.

\section{CFVBench}

\subsection{Preliminary}
Video-based MRAG is typically formulated as a two-stage process: 
(1) \textbf{Retrieval.} Given a textual question $T_q$, a retriever $R$ selects a subset of videos (or video segments) 
$V = \{V_{1}, V_{2}, \dots, V_{k}\}$
that are most relevant to the question. The goal of this stage is to identify a minimal set of videos containing sufficient information to answer question $T_q$, thereby enabling efficient downstream processing.   
(2) \textbf{Generation.} A MLLM $M$ then takes the question $T_q$ and the retrieved video subset $V$ as input, i.e., $(T_q, V)$, and synthesizes information across multiple modalities and temporal video segments to generate the final answer $A$. Formally, this can be expressed as
\begin{equation}
A = M(T_q, V)
\label{eq:vlm_answer}
\end{equation}


To evaluate retrieval and generation while assessing fine-grained multimodal comprehension ability, we introduce CFVBench, a large-scale, manually verified benchmark of high-density videos.

\subsection{Dataset Construction}

As shown in Fig.~\ref{fig:pipeline}, the construction of CFVBench follows a 4-stage pipeline designed to ensure diversity, difficulty, and quality: 

\textbf{(1) Video Curation.}
We curated videos from YouTube spanning diverse domains such as finance, climate, and environment. Our focus was on content that inherently requires multimodal and fine-grained reasoning. The videos fall into three main categories:
(i) \textit{Structured-Data Videos}: Videos containing dense visual information such as charts and tables, which require accurate extraction of fine-grained details and numerical reasoning, aligned with transcripts.
(ii) \textit{Tutorials}: Instructional content (e.g., software or websites utilizing workflows) where UI operations are tightly coupled with audio scripts, demanding cross-modal reasoning.
(iii) \textit{News}: News reports that integrate live footage, scrolling captions, and occasional charts or texts, requiring models to capture objects, entities, scenes, dynamics, and transient information.

The specific websites corresponding to these categories are listed in Appendix~\ref{appendix:source}. Each curated video is paired with its transcript and audio track, ensuring that CFVBench offers comprehensive multimodal coverage across video, text, and audio, as well as diverse format (such as charts and tables) in the video frames.

\textbf{(2) Keypoints Generation.}  
First, we used Qwen2.5-VL-72B-Instruct~\cite{qwen2.5-VL} to automatically identify informative timestamps under a category-specific protocol: For tutorial videos, the model segmented steps and annotated each operation concisely. For structured data videos, it detects comprehensive data formats such as charts, tables, maps, and flowcharts, and outputs start and end times with formatted semantic tags in JSON. For news videos, salient frames were recognized and described at the image level (e.g., polar bears walking on ice). All outputs were stored in a unified JSON schema with temporal boundaries and semantic annotations, providing reliable anchors for keypoint extraction.


Following~\cite{zhu-etal-2025-rageval, yang2025benchmarking}, keypoints are fine-grained, semantically independent facts from videos or subtitles, serving as ground truth for QA construction and evaluation. To handle diverse video domains, we designed tailored extraction protocols. For {Structured-Data} and {News} videos, we extracted (i) {textual keypoints} from subtitles via DeepSeek-R1~\cite{deepseekai2025deepseekr1incentivizingreasoningcapability}, and (ii) {visual keypoints} from frames aligned with timestamps using Qwen2.5-VL-72B-Instruct. For {Tutorial} videos, where narration and on-screen actions are tightly coupled, we generated {hybrid keypoints} from subtitles, frames, and timestamps, encoding explicit (action, object, outcome) triples (e.g., “the user clicks the ‘File’ button → the ‘Save As’ dialog box opens → a file path is displayed”). This framework ensures QA tasks require cross-modal reasoning rather than a single modality. To improve accuracy, we applied the LLMs three times on each sample and took the intersection of the results.

\begin{figure}[t!]
    \centering
    \includegraphics[width=0.8\linewidth]{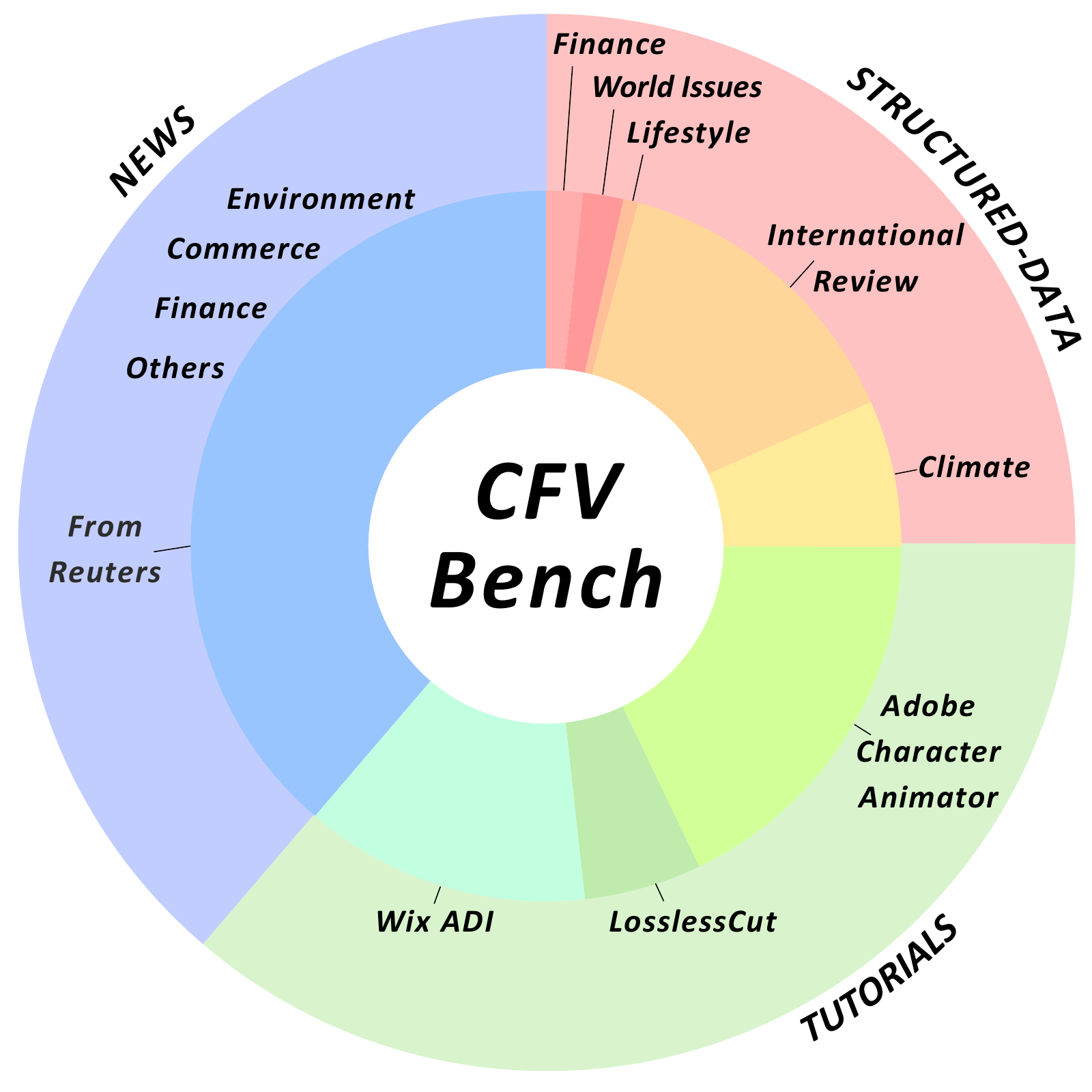}
    \caption{The distribution of videos in CFVBench.} 
    \label{fig:type}
\end{figure}

\textbf{(3) QA Pair Generation.}  
Building on extracted keypoints, we constructed open-ended QA pairs with DeepSeek-R1. We generate (i) \textit{single-hop} questions, answerable from a single keypoint (e.g., factual or numerical detail) from a single modality, and (ii) \textit{multi-hop} questions, requiring integrating two or more key points (e.g., sequential tutorial steps or combining visual and textual evidence) from multiple modalities. Questions primarily follow factual and procedural forms (“What,” “How”) while also including causal (“Why”), comparative (“How does … compare”), and conditional (“What happens if …”) patterns. Answers are grounded in keypoints without external inference, QA are independent and align with the general cognitive level of human video viewing, supporting reliable evaluation of retrieval and multi-step reasoning.

\textbf{(4) QA Pair Filtering.}  
To ensure the quality of generated QA pairs, we employed DeepSeek-R1 to make each QA pair automatically checked against 4 criteria: (i) for multi-hop questions, all listed keypoints must be necessary to construct the complete answer; redundant or unused keypoints were removed, (ii) duplicates across QA pairs were eliminated, (iii) QA pairs unrelated to the topic were discarded, and (iv) overly mechanical or unnatural questions (e.g., over-detailing or forced associations) were filtered out.

To further enhance the quality of CFVBench and ensure questions are meaningful to humans and align with real-life application scenarios, we conduct human modification. We recruited 13 graduate students from local universities, all with academic backgrounds in computer science, journalism, or English, and IELTS scores above 7.0. Before annotation, they underwent training that introduced the task objectives and detailed guidelines (aligned with QA pair filtering criteria), demonstrated examples of both correct and incorrect annotations, emphasized data privacy and responsible usage, and clarified that they need to modify the QA pairs to make semantics more precise while keeping content unchanged. During annotation, each annotator was assigned a subset of QA pairs, with 30\% overlapping between two annotators. Annotators were also instructed to adjust temporal segments when initial video timestamps were incorrectly split. A QA pair was finalized only when both annotators provided consistent annotations, ensuring dataset reliability. Additionally, for quality control, 2 supervising annotators randomly sampled 20\% of every 100 QA pairs. If any issue was found, the process was repeated until all sampled pairs satisfied the standards.


The prompts in this section are shown in Appendix~\ref{appendix:prompt} (1-9).

\subsection{Dataset Composition}

After the dataset construction process, CFVBench consists of 3 core video categories: structured-data videos, tutorials, and news. as shown in Fig.~\ref{fig:type}, these categories are further divided into topics such as finance, climate, software operations, and commerce, covering key areas of real-world applications. Statistically, as reported in Table~\ref{tab:dataset-stats} of Appendix~\ref{appendix: Statistics}, CFVBench contains a total of 5,360 question-answer pairs. To evaluate different levels of reasoning complexity, the test set includes 3,703 single-hop questions and 1,660 multi-hop questions. On average, multi-hop questions require integrating 2.72 keypoints across different modalities, placing higher demands on the model's reasoning capabilities. Additionally, the average video length is 232.3 seconds, and the average question length is 15.22 words, reflecting rich content and moderate question complexity.




\begin{table*}[t]
    \caption{Retrieval methods performance comparison under different scenarios. The top 3 best scores for each metric are highlighted as \bfirst{}{first}, \bsecond{}{second}, and \bthird{}{third}.}
    \label{tab:performance_full}
    \centering
    \small
    \setlength{\tabcolsep}{3pt}
    \resizebox{0.95 \textwidth}{!}{
    \begin{tabular}{c|cccc|cccc|cccc|cccc}  
        \toprule
        \centering\raisebox{-4ex}[0pt][0pt]{Model} & \multicolumn{4}{c|}{\raisebox{-2ex}[0pt][0pt]{Overall}} & \multicolumn{4}{c|}{\raisebox{-2ex}[0pt][0pt]{Multi-Point}} & \multicolumn{8}{c}{\raisebox{0ex}[0pt][0pt]{Single-Point}} \\
        \cmidrule(l){10-17}
        & \multicolumn{4}{c|}{} & \multicolumn{4}{c|}{} & \multicolumn{4}{c|}{Text-Single-Point} & \multicolumn{4}{c}{Video-Single-Point} \\
        \cmidrule(lr){2-5} \cmidrule(lr){6-9} \cmidrule(lr){10-13} \cmidrule(l){14-17}
        & R@1 & R@3 & R@5 & R@10 & R@1 & R@3 & R@5 & R@10 & R@1 & R@3 & R@5 & R@10 & R@1 & R@3 & R@5 & R@10 \\
        \midrule
        Nomic-embed-text & 51.01 & 64.53 & 69.42 & 76.11 & 41.74 & 56.20 & 60.84 & \third{68.19} & 58.33 & 71.01 & 75.74 & 82.13 & 49.26 & 63.12 & 68.62 & 75.05 \\
        \midrule
        Imagebind & 37.16 & 54.03 & 61.21 & 70.63 & 25.42 & 38.97 & 45.18 & 54.91 & 42.12 & 61.23 & 68.82 & 79.97 & 42.98 & 59.95 & 67.62 & 74.67 \\
        Internvideo & 30.05 & 45.12 & 52.32 & 63.21 & 20.00 & 29.27 & 35.36 & 43.73 & 34.16 & 52.03 & 59.70 & 71.97 & 35.32 & 52.59 & 60.34 & 71.88 \\
        Languagebind & 46.18 & 61.88 & 68.13 & 76.17 & 32.16 & 43.37 & 48.67 & 57.28 & 53.93 & 71.84 & 78.48 & \third{86.31} & 49.72 & 67.07 & \second{73.81} & \second{81.48} \\
        \midrule
        Imagebind + nomic-embed-text & \second{55.24} & \second{70.05} & \second{75.64} & \second{81.33} & \second{43.49} & \first{58.25} & \first{63.79} & \first{71.62} & \second{63.14} & \second{77.81} & \second{83.33} & \second{88.05} & \second{55.61} & \second{70.72} & 76.52 & \third{81.25} \\
        Internvideo + nomic-embed-text & \third{54.09} & \third{67.55} & \third{72.83} & \third{78.76} & \third{42.83} & \third{56.74} & \second{62.65} & \second{69.06} & \third{61.69} & \third{74.83} & \third{79.35} & 85.15 & \third{54.37} & \third{67.85} & \third{73.74} & 79.24 \\
        Languagebind + nomic-embed-text & \first{57.51} & \first{71.47} & \first{76.31} & \first{81.37} & \first{45.78} & \second{57.71} & \third{62.59} & 68.13 & \first{65.46} & \first{79.85} & \first{84.78} & \first{89.38} & \first{57.70} & \first{73.50} & \first{78.07} & \first{83.42} \\
        \bottomrule
    \end{tabular}
    }
\end{table*}

\section{Evaluation on CFVBench}
MRAG frameworks are typically composed of two core stages: retrieval and generation. To thoroughly assess existing video-based MRAG methods' performance, we separately evaluate both stages of MRAG on the proposed CFVBench. The following sections will first introduce the baselines, evaluation metrics, and implementation details relevant to both stages, followed by a comprehensive analysis of the experimental results.

\begin{table*}[t]
    \caption{Generation performance comparison of different MLLMs on CFVBench. The top 3 best scores for each metric are highlighted as \bfirst{}{first}, \bsecond{}{second}, and \bthird{}{third}. (\rfirst{}{red} and \bfirst{}{blue} to distinguish open-source and close-source models).}
    \label{tab:eval_results}
    \centering
  \small
    \resizebox{0.95 \textwidth}{!}{%
  \begin{tabular}{*{12}{c}}
  \toprule
        Model & $Recall_{t}$ & $Recall_{v}$ & $Recall_{all}$ & Precision & F1 & Rouge$\_$L & St$\_$cos & Likert & Fact$\_$cov & Vis$\_$use & Ling$\_$prec \\
  \midrule
        claude-opus-4              & \redthird{0.2031} & \redsecond{0.2162} & \redsecond{0.2079} & \redsecond{0.1667} & \redsecond{0.1850} & \redthird{0.0909} & \redthird{0.6545} & \redthird{3.2000} & \redthird{1.8909} & \redsecond{2.3818} & \redthird{3.1273} \\
        gpt-5-chat         & \redsecond{0.2079} & \redthird{0.1552} & \redthird{0.1887} & \redthird{0.1220} & \redthird{0.1482} & \redsecond{0.1477} & \redsecond{0.6705} & \redsecond{3.5455} & \redfirst{2.2386} & \redthird{2.3409} & \redsecond{3.7045} \\
        gemini-2.5-flash            & \redfirst{0.2843} & \redfirst{0.3051} & \redfirst{0.2919} & \redfirst{0.1673} & \redfirst{0.2127} & \redfirst{0.1778} & \redfirst{0.8000} & \redfirst{3.6889} & \redsecond{2.1000} & \redfirst{2.5333} & \redfirst{3.9444} \\
  \midrule
        Gemma-3-12b-it   & 0.0685 & 0.0601 & 0.0656 & 0.0837 & 0.0736 & 0.1287 & 0.6045 & 2.8205 & 1.4817 & 2.0272 & 3.0366 \\
        Qwen2.5-VL-7B-Instruct     & 0.1152 & 0.1220 & 0.1176 & 0.1067 & 0.1119 & 0.1376 & 0.6223 & 3.0996 & 1.6807 & 2.0953 & 3.7785 \\
        InternVL-3.5-8B            & 0.1805 & 0.2188 & 0.1936 & \third{0.1440} & 0.1652 & 0.1601 & 0.6814 & 3.4934 & 2.1734 & 2.3189 & 4.1022 \\
        llava-llama-3-8b                      & 0.1583 & \second{0.2810} & 0.2018 & 0.0985 & 0.1324 & \first{0.2209} & \second{0.7933} & 3.5095 & 2.1308 & 2.2473 & 3.8312 \\
        MiniCPM-V-2$\_$6                    & \third{0.2407} & \first{0.3793} & \first{0.2892} & 0.1424 & \second{0.1908} & 0.1702 & \third{0.7766} & 3.6383 & 2.4286 & 2.3736 & 3.9560 \\
        InternVL3$\_$5-14B                 & 0.1858 & 0.2354 & 0.2034 & \second{0.1470} & \third{0.1707} & \third{0.2017} & 0.7272 & 3.7070 & 2.2843 & 2.3807 & 4.1494 \\
        Mistral-Small-3.2-24B-Instruct     & 0.1883 & 0.2041 & 0.1931 & 0.1069 & 0.1376 & 0.1639 & 0.6995 & 3.7104 & 2.2623 & 2.3880 & 4.0273 \\
        InternVL3$\_$5-30B        & 0.2156 & 0.1818 & 0.2046 & 0.1193 & 0.1507 & 0.2000 & 0.7444 & 3.8296 & \third{2.4556} & \third{2.4000} & 4.1148 \\
        Intern-S1-mini                    & \first{0.2941} & \third{0.2353} & \second{0.2745} & \first{0.1609} & \first{0.2029} & \third{0.2017} & 0.7647 & \third{3.8319} & 2.2797 & 2.3814 & \first{4.4153} \\
        Magistral-Small                   & 0.2303 & 0.1512 & 0.2045 & 0.1057 & 0.1394 & 0.1883 & 0.7324 & \second{3.9351}& \first{2.5490} & \second{2.5621} & \third{4.1569} \\
        Gemma-3-27b                    & \second{0.2655} & 0.1935 & \third{0.2400} & 0.1180 & 0.1582 & \second{0.2079} & \first{0.8020} & \first{3.9703} & {\second{2.4851}} & \first{2.5644} & \second{4.2277} \\
        \bottomrule
    \end{tabular}
    }
\end{table*}

\subsection{Evaluation Settings}
\textbf{Baselines.}
The retrieval stage employs 3 widely-utilized multimodal retrieval methods, including ImageBind~\cite{girdhar2023imagebind}, InternVideo~\cite{wang2023internvid}, and LanguageBind~\cite{zhu2023languagebind} that align video and text representations within a shared embedding space. Building upon this, the nomic-embed-text~\cite{nussbaum2024nomic} model is incorporated to specifically capture semantic similarities in the textual modality. 
After the retrieval stage, the top-$k$ video clips identified by the best-performing retrieval method are passed to the generation stage.



During the generation stage, we combine the retrieval results and conduct a zero-shot evaluation on CFVBench with 14 MLLMs, including: Gemma-3-12b-it~\cite{gemma_2025}, Gemma-3-27b-it, Qwen2.5-VL-7B-Instruct~\cite{qwen2.5-VL}, InternVL-3.5-8B~\cite{wang2025internvl3_5}, InternVL3\_5-14B, InternVL3\_5-30B-A3B, llava-llama-3-8b-v1\_1~\cite{2023xtuner}, Intern-S1-mini~\cite{bai2025intern}, MiniCPM-V-2\_6~\cite{yao2024minicpm}, Magistral-Small-2509~\cite{rastogi2025magistral}, Mistral-Small-3.2-24B-Instruct-2506, and 3 close-source models: gpt-5-chat-latest~\cite{gpt-5}, gemini-2.5-flash-lite-preview-09-2025~\cite{gemini-2-5-flash}, claude-opus-4-20250514~\cite{claude-4}.


\textbf{Evaluation Metrics.} 
For retrieval evaluation, we adopt Recall@K~\cite{yang2025benchmarking} (R@K), which measures whether at least one relevant video is included among the top-$k$ retrieved results. We report Recall@1, Recall@3, Recall@5, and Recall@10 to capture performance across different cutoff thresholds. We evaluate two query types: (1) multi-point: queries require retrieving videos containing multiple keypoints across different modalities; (2) single-point: queries target videos with a single keypoint, further divided into text-single-point (speech/transcript-based, where keypoints are derived from audio transcripts) and video-single-point (visual content-based, where keypoints originate from visual information) to assess modality-specific retrieval capabilities.

In the generation stage, following~\cite{yang2025benchmarking, zhu-etal-2025-rageval}, evaluation metrics are divided into two categories.
(1) \textit{Automatic evaluation}: To assess a model’s ability to leverage multimodal information in CFVBench, we adopt keypoint-based recall, precision, and F1-score~\cite{powers2020evaluation}. Recall is computed for video ($Recall_{v} = C_v/N_v$) and textual keypoints ($Recall_{t} = C_t/N_t$), where $C_v$, $C_t$ denote correctly matched keypoints and $N_v$, $N_t$ mean total keypoints, the overall recall is $Recall_{all} = (C_v + C_t)/(N_v + N_t)$. Precision measures the proportion of correctly covered keypoints among all factual claims ($M$) in the response ($Precision = (C_v + C_t)/M$), and F1 is their harmonic mean. We also introduce ROUGE-L~\cite{lin2004rouge} to capture lexical overlap.
(2) \textit{LLM-as-Judge evaluation}: We average the results from Qwen3-8B-Instruct~\cite{qwen3technicalreport} and GLM-4-9B~\cite{glm2024chatglmfamilylargelanguage} to holistically score model outputs. Each response is rated on three dimensions: Factual Coverage (Fact\_cov), Visual Detail Usage (Vis\_use), Linguistic Precision (Ling\_prec), and assigned a final score on a 1–5 Likert scale~\cite{joshi2015likert}. Additionally, we compute semantic similarity (St\_cos) between generated and reference answers using all-MiniLM-L6-v2~\cite{reimers-2019-sentence-bert}. Relevant prompts are in Appendix~\ref{appendix:prompt} (1-9).

\textbf{Implementation Details.} 
During the retrieval stage, we follow the procedure in~\cite{VideoRAG}, which leverages both audio transcriptions and video captions. Specifically, the audio of each video is transcribed using Faster-Distil-Whisper-Large-v3~\cite{gandhi2023distilwhisper}, while visual frames sampled at 6-second intervals are captioned with MiniCPM-V-4-int4~\cite{yao2024minicpm}, conditioned on the corresponding audio information. Based on these modalities, we generate multimodal embeddings from the raw videos and extract text embeddings from the generated descriptions. The top-$k$ full videos are first retrieved using cosine similarity, after which each is segmented into 30-second clips, and the results of best-performing retrieval method are then passed to the generation stage.



In the generation stage, we follow~\cite{VideoRAG} and further segment each video into 6-second clips, from which the five most relevant ones are selected. For each clip, its transcript is combined with 5 frames sampled at 6-second intervals, and the resulting multimodal input is fed into the MLLM in the retrieval order to generate the final answer. To ensure deterministic outputs, we set the temperature to 0.1 and top-$p$ to 1. All MLLMs are evaluated using identical prompts Appendix ~\ref{appendix:prompt} (11) on NVIDIA A100 GPUs with 80GB of memory.



\vspace{-10pt}
\subsection{Evaluation Results}
\label{sec:eval}
\newcommand{\best}[1]{\cellcolor{blue!20}#1}  

\textbf{Analysis of Retrieval Results.} 
As shown in Table~\ref{tab:performance_full}, we could find: combining textual and multimodal embeddings yields consistent improvements, with {LanguageBind + nomic-embed-text} achieving the best overall $\mathrm{R}@10$ of $81.37\%$, underscoring the complementary strengths of text and video representations. In comparison, multimodal embeddings alone generally outperform text-only retrieval on single-point queries, where the best video model (LanguageBind) achieves $76.17\%$ $\mathrm{R}@10$, slightly above the text baseline ($76.11\%$). However, multi-point queries remain particularly challenging. An interesting observation is that multimodal embeddings underperform text-only embeddings under multi-point setting, suggesting that the video modality may introduce noise and hinder multi-hop reasoning. Moreover, across all multimodal methods, performance in multi-point setting lags behind single-point results, with even the best hybrid approach reaching only $71.62\%$ $\mathrm{R}@10$. This substantial drop highlights that current models still struggle with fine-grained temporal integration and multi-hop reasoning.

\begin{figure}[t!]
    \centering
    \includegraphics[width=0.9\linewidth]{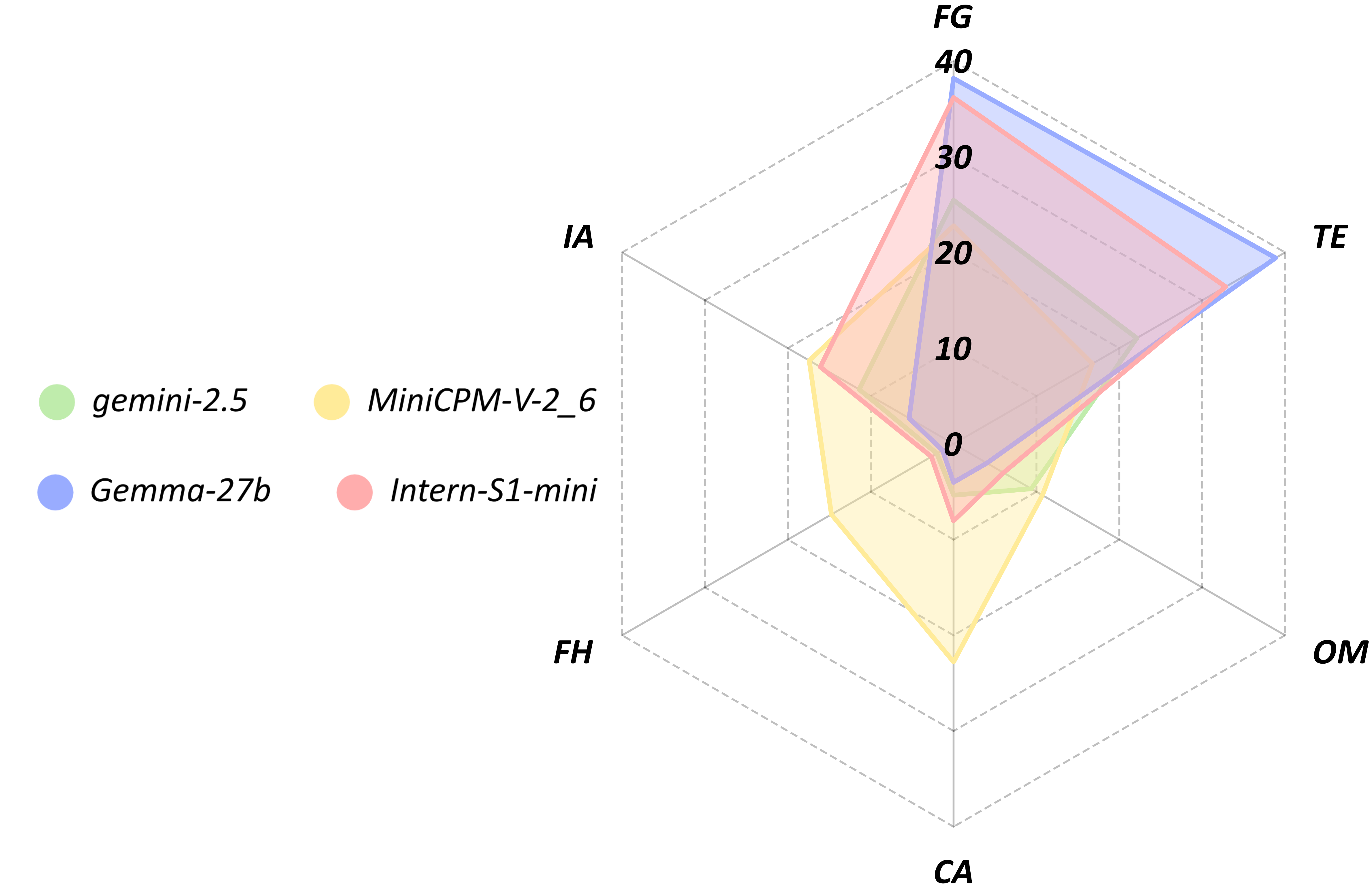}
    \caption{Human evaluation of typical MLLMs on CFVBench.}
    \label{fig:human_value_raw}
\end{figure}

\textbf{Analysis of Generation Results.}\label{Chap:QA results}
As shown in Table~\ref{tab:eval_results}, the experimental results on CFVBench reveal several insights: (1) there are substantial performance differences among different MLLMs. For example, considering $Recall_t$, the highest-performing model achieves 0.2941 while the lowest is only 0.0685, indicating a large gap in the ability to extract key fine-grained information. (2) Although closed-source models that are not specifically designed for video tasks, such as gemini, do not show significant advantages over newer open-source video-oriented models, they demonstrate relatively stable performance across metrics, highlighting their effectiveness and reliability in video-based MRAG task. (3) The overall performance of all evaluated models remains modest across automatic and LLM-as-Judge evaluation, demonstrating the challenge posed by CFVBench and underscoring the necessity and importance of studying fine-grained video-based MRAG.

To further investigate model behavior, we conducted a manual error analysis. We randomly sampled 100 generated answers from Gemma-3-27B, MiniCPM-V-2\_6, Gemma-27B, and Intern-S1-mini, and asked 3 annotators to label each response according to 6 predefined error types. Majority voting was applied, and Fleiss’s kappa~\cite{fleiss1973equivalence} of 0.81 indicated strong inter-annotator agreement. The evaluated error categories include Fine-Grained Detail Omission (FG), Transient Event Neglect (TE), Object/Entity Misidentification (OM), Contextual Association Error (CA), Factual Hallucination (FH), and Incomplete Answer (IA). As shown in Fig.~\ref{fig:human_value_raw}, we observe: (1) all models frequently suffer from FG and TE errors, with Gemma-27B and Intern-S1-mini being most affected, highlighting the challenge of fine-grained information capture; (2) MiniCPM-V-2\_6 produces fewer FG and TE errors but more CA, FH, and IA cases, suggesting that although it excels at detail perception, it is more vulnerable to hallucination and reasoning instability; (3) Gemini-2.5 demonstrates the most balanced performance, showing stronger robustness in the fine-grained video-based MRAG task.

\begin{figure}
    \centering
    \includegraphics[width=1\linewidth]{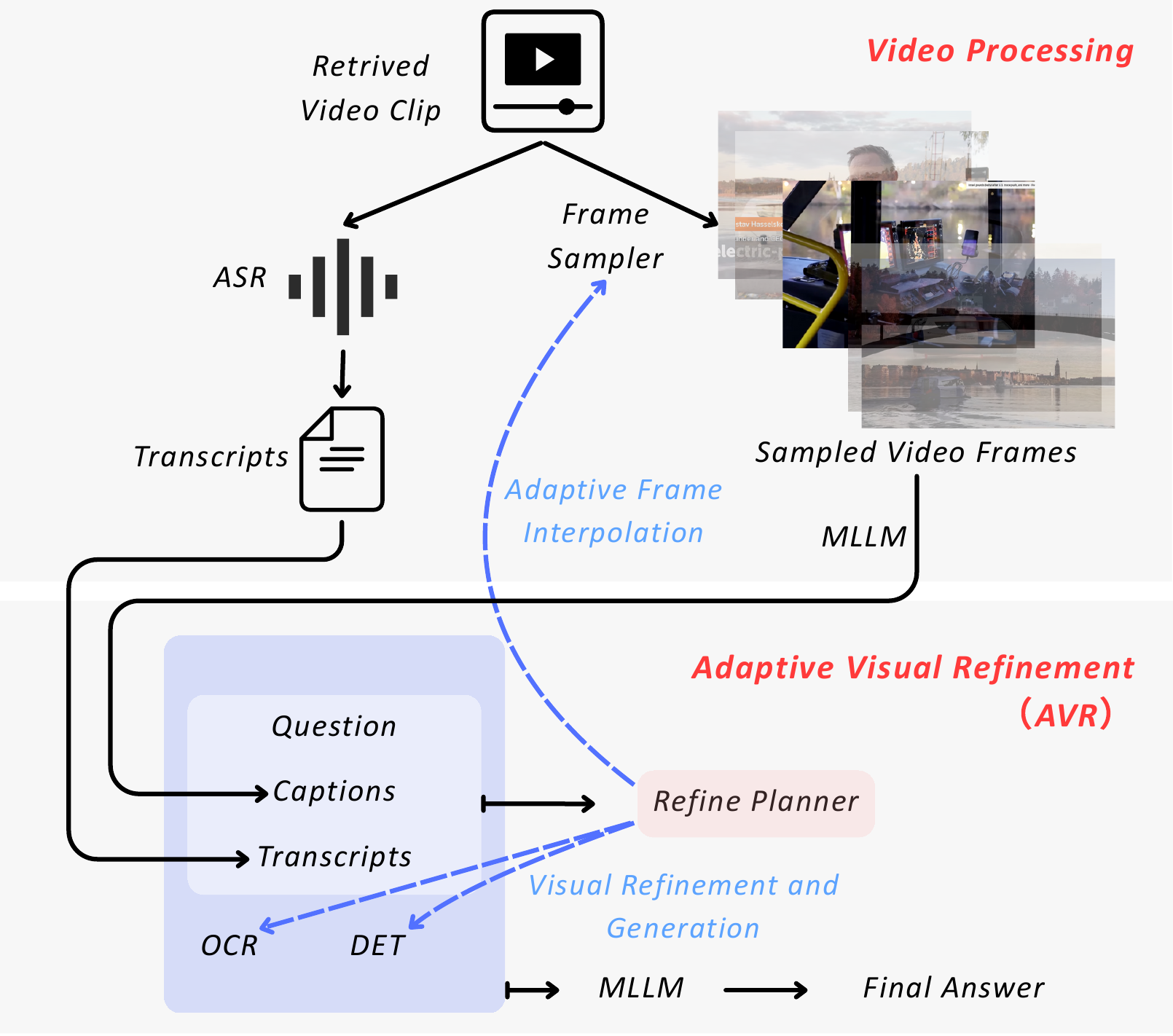}
    \caption{The workflow of the AVR framework.}
    \label{fig:placeholder}
\end{figure}

\section{Methodology}
To address the challenge of fine-grained comprehension, we propose Adaptive Visual Refinement (AVR) framework. First, {adaptive frame interpolation} mechanism assesses evidence from sparse frame sampling and adaptively increases frame density for critical segments. Second, {visual refinement and generation} strategy performs on-demand tool invocation, such as OCR~\cite{vedhaviyassh2022comparative} and object detection (DET)~\cite{cheng2024yolo}, and aggregates multimodal evidence to generate the final answer. The workflow of AVR is shown in Fig.~\ref{fig:placeholder}.


\begin{table*}[t]
    \caption{Generation performance comparison on CFVBench with the proposed AVR framework (marked with *).}
    \label{tab:eval_results_avr}
    \centering
  \small
    \resizebox{0.95 \textwidth}{!}{%
  \begin{tabular}{*{12}{c}}
  \toprule
        Model & $Recall_{t}$ & $Recall_{v}$ & $Recall_{all}$ & Precision & F1 & Rouge$\_$L & St$\_$cos & Likert & Fact$\_$cov & Vis$\_$use & Ling$\_$prec \\
  \midrule
        claude-opus-4*              & \redfirst{0.3125} & \redsecond{0.3077} & \redfirst{0.3103} & \redfirst{0.2000} & \redfirst{0.2432} & \redthird{0.1765} & \redthird{0.7059} & \redthird{3.6471} & \redsecond{2.4706} & \redfirst{2.8235} & \redthird{3.8824} \\
        gpt-5-chat*                 & \redsecond{0.2946} & \redthird{0.2727} & \redthird{0.2874} & \redthird{0.1644} & \redthird{0.2092} & \redfirst{0.2211} & \redsecond{0.7263} & \redfirst{4.1158} & \redfirst{2.8526} & \redsecond{2.6632} & \redfirst{4.0947} \\
        gemini-2.5-flash*              & \redthird{0.2210} & \redfirst{0.4625} & \redsecond{0.2950} & \redsecond{0.1778} & \redsecond{0.2219} & \redsecond{0.1986} & \redfirst{0.8582} & \redsecond{3.7801} & \redthird{2.3901} & \redthird{2.5887} & \redsecond{3.9929} \\
  \midrule
        Gemma-3-12b-it*       & 0.0621 & 0.0730 & 0.0660 & 0.1013 & 0.0799 & 0.1354 & 0.6247 & 3.0143 & 1.5879 & 2.1817 & 3.1508 \\
        Qwen2.5-VL-7B-Instruct* & 0.1069 & 0.1449 & 0.1204 & 0.1049 & 0.1121 & 0.1513 & 0.6454 & 3.2943 & 1.6980 & 2.1510 & 3.9584 \\
        InternVL-3.5-8B*       & 0.1638 & 0.2941 & 0.2051 & 0.1457 & 0.1704 & 0.1667 & 0.7583 & 3.5542 & 2.2000 & 2.3958 & 4.1208 \\
        llava-llama-3-8b*      & 0.1523 & \third{0.3321} & 0.2160 & 0.1116 & 0.1472 & 0.2328 & 0.8017 & 3.5701 & 2.1490 & 2.3552 & 3.9380 \\
        MiniCPM-V-2\_6*        & 0.2178 & \first{0.4561} & \second{0.3038} & \second{0.1846} & \second{0.2297} & 0.2111 & 0.8111 & 3.7111 & 2.4778 & 2.4333 & 4.0222 \\
        InternVL3\_5-14B*      & 0.1692 & 0.2839 & 0.2099 & \third{0.1600} & \third{0.1816} & 0.2127 & 0.7485 & 3.8613 & 2.3691 & 2.4914 & \third{4.2580} \\
        Mistral-Small*         & 0.1857 & 0.2451 & 0.2035 & 0.1090 & 0.1420 & 0.1799 & 0.7143 & 3.7937 & 2.2910 & 2.4762 & 4.0952 \\
        InternVL3\_5-30B*      & \third{0.2239} & 0.2941 & 0.2475 & 0.1217 & 0.1632 & \first{0.2712} & \third{0.8136} & \third{3.9831} & \third{2.5085} & \third{2.5339} & 4.1525 \\
        Intern-S1-mini*         & \first{0.2946} & \second{0.3667} & \first{0.3198} & \first{0.1993} & \first{0.2456} & \second{0.2626} & \second{0.8182} & 3.9495 & 2.3232 & 2.4343 & \first{4.4646} \\
        Magistral-Small*        & 0.2143 & 0.2958 & 0.2400 & 0.1176 & 0.1579 & \third{0.2598} & 0.7480 & \second{4.0000} & \first{2.6220} & \second{2.6220} & 4.2126 \\
        Gemma-3-27b*           & \second{0.2768} & 0.2581 & \third{0.2701} & 0.1247 & 0.1706 & 0.2500 & \first{0.8500} & \first{4.1000} & \second{2.5100} & \first{2.7800} & \second{4.2600} \\
        \bottomrule
    \end{tabular}
    }
\end{table*}

\subsection{Adaptive Frame Interpolation}
Adaptive frame interpolation evaluates the sufficiency of initial evidence and determines whether denser sampling is needed. For each retrieved video segment $V_i$, we first conduct video processing:
(1) \textit{Sparse Frame Sampling}: extract $N_{sparse}$ frames (e.g., $N_{sparse}=5$) to form a sparse frame set $F_{sparse}$.
(2) \textit{Audio Transcription}: generate subtitles $T_{asr}$ using Automatic Speech Recognition (ASR) model Whisper~\cite{gandhi2023distilwhisper}.
(3) \textit{Preliminary Summary}: obtain an initial summary $C_{init}$ from $F_{sparse}$ using the MLLM to be evaluated.

We then feed the query $T_q$, subtitles $T_{asr}$, and summary $C_{init}$ into the MLLM-based refinement planner with a refinement prompt (Appendix~\ref{appendix:prompt} (10)), which explicitly instructs the model to assess whether the current evidence suffices to answer the query and to identify potentially missing information. The planner outputs an \textit{Information Sufficiency Score} $S \in [0,10]$, which integrates an overall \textit{answerability score} measuring how well the query can be addressed given the current evidence and an \textit{information density score} evaluating the richness and granularity of visual content in $F_{sparse}$ and $T_{asr}$. A higher score $S$ indicates sufficient evidence, while a lower score signals missing fine-grained details. 

Based on the information sufficiency score $S$, AVR determines the number of frames $N_{target}$ to sample according to:
\begin{equation}
N_{target} = f(S) = \begin{cases}
N_{sparse} & \text{if } S > \theta \\
 N_{min} + (N_{max} - N_{min}) \cdot \frac{\theta - S}{\theta}  & \text{if } S \le \theta
\end{cases}
\end{equation}
where $\theta$ is the threshold (set to 5.0) and $[N_{min}, N_{max}]$ is the refinement interval (set to $[20,40]$; following \citet{kandhare2024empirical} and \citet{luo2024video}, they both identify that the sampling rate achieves optimal results at around 1 fps). If $S>\theta$, the original sparse frames are retained; otherwise, denser sampling is applied proportionally to the severity of the information gap. Interpolated frames are filtered by visual similarity to reduce redundancy, and only diverse frames are retained for caption generation. Captions from these refined frames are merged with existing subtitles to construct a richer, fine-grained representation, which is subsequently fed into the visual refinement and generation stage of AVR.

\subsection{Visual Refinement and Generation}
Guided by the MLLM-based refinement planner, we perform on-demand tool invocation and multimodal evidence aggregation to construct a rich feature representation for final answer generation.

\textbf{(1) On-demand Tool Invocation.}
The refinement planner determines whether specialized tools are required (Appendix~\ref{appendix:prompt} (10)). Specifically, OCR and DET tools are invoked only when explicitly required: OCR is used when the query depends on reading symbolic or structured visual data (e.g., numbers, UI elements, tables, charts), while DET is applied when answering requires recognizing entity categories, attributes, or spatial relations (e.g., object state, interactions, fine-grained classification).

If OCR is triggered, we apply EasyOCR~\cite{vedhaviyassh2022comparative} to the selected target frames $F_{target}$ to extract symbolic or numeric information, such as textual overlays, tables, or chart values, and aggregate the results into $T_{ocr}$. If object detection (DET) is requested, we derive a task-specific detection vocabulary from the query $T_q$ using a prompt-based keyword generator (Appendix~\ref{appendix:prompt} (13)) and run YOLO-World~\cite{Cheng2024YOLOWorld} to detect corresponding entities in $F_{target}$. The detection results (object class and spatial attributes) are converted into textual descriptions $T_{det}$.

\textbf{(2) Evidence Aggregation and Final Generation.}
For the selected target frames $F_{target}$, we employ an MLLM-based captioner to generate a detailed video summary $C_{target}$. All available evidence sources are then concatenated to form the enriched feature set ${F}_{rich} = [{T_q, T_{asr}, C_{target}, T_{ocr}, T_{det}}]$, where $T_{ocr}$ and $T_{det}$ are optional. Finally, the MLLM integrates ${F}_{rich}$ to generate the final answer $A = {M}(T_q, {F}_{rich})$.

All MLLM-based components in AVR (including the refinement planner, captioner, and answer generator) employ the same MLLM that is being evaluated, ensuring consistency across stages and eliminating potential confounding effects from external models.

\begin{table}
\caption{Ablation study on AVR, with \textcolor{softgreen}{green} values showing the change in indicator scores relative to results w/o AVR.}
\label{tab:ablation_avr}
\centering
\resizebox{0.46 \textwidth}{!}{  
\begin{tabular}{c c c}
\toprule
Model & $Recall_{v}$ & Vis$\_${use} \\
\midrule
Gemma-3-27b & 0.1935  & 2.5644 \\
Gemma-3-27b (+ frames) & 0.2222 \textcolor{softgreen}{+14.83\%} & 2.7059 \textcolor{softgreen}{+5.52\%} \\
Gemma-3-27b (+ frames \& DET) & 0.2407 \textcolor{softgreen}{+24.39\%} & 2.7326 \textcolor{softgreen}{+6.56\%} \\
Gemma-3-27b (+ frames \& OCR) & 0.2545 \textcolor{softgreen}{+31.52\%} & 2.7529 \textcolor{softgreen}{+7.35\%} \\
  \midrule
InternVL3\_5-30B & 0.1818 & 2.4000  \\
InternVL3\_5-30B (+ frames) & 0.2593 \textcolor{softgreen}{+42.63\%} & 2.4824 \textcolor{softgreen}{+3.43\%} \\
InternVL3\_5-30B (+ frames \& DET) & 0.2778 \textcolor{softgreen}{+52.81\%} & 2.5059 \textcolor{softgreen}{+4.41\%} \\
InternVL3\_5-30B (+ frames \& OCR) & 0.2647 \textcolor{softgreen}{+45.60\%} & 2.5085 \textcolor{softgreen}{+4.52\%} \\
\bottomrule
\end{tabular}
}
\end{table}

\section{Experiment with AVR}

\textbf{Results with AVR.} 
Table~\ref{tab:eval_results_avr} demonstrates the effectiveness of AVR. Incorporating AVR leads to consistent metric improvements across most MLLMs, with particularly notable gains in $Recall_v$; for instance, MiniCPM-V-2\_6 rises from 0.3793 to 0.4561, confirming AVR’s ability to enhance fine-grained multimodal comprehension through dynamic frame sampling and selective OCR/object detection. These results suggest that AVR enables models to better capture critical visual details and reason over dense video content. A slight drop in $Recall_t$ for some models reflects the trade-off between richer visual coverage and potential retrieval noise.
We also present performance and robustness analyses, including method comparisons and retrieval reordering experiments. Detailed results are available in Appendix~\ref{appendix:method} and Appendix~\ref{appendix:Robustness}.

\begin{figure}
    \centering
    \includegraphics[width=1\linewidth]{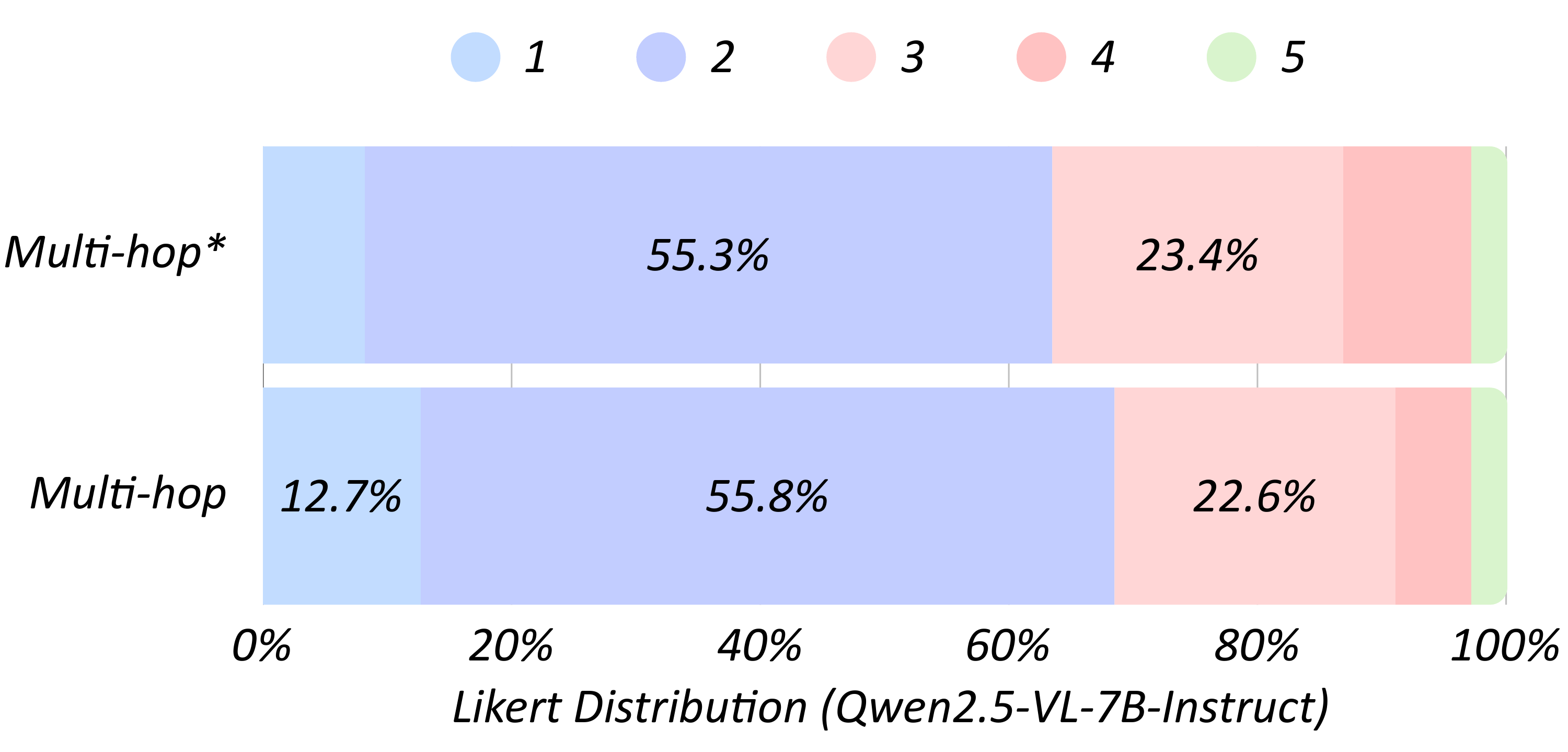}
    \caption{Performance on multi-hop questions, with (*) indicating use of AVR.}
    \label{fig:hop}
\end{figure}

\begin{figure}
    \centering
    \includegraphics[width=1\linewidth]{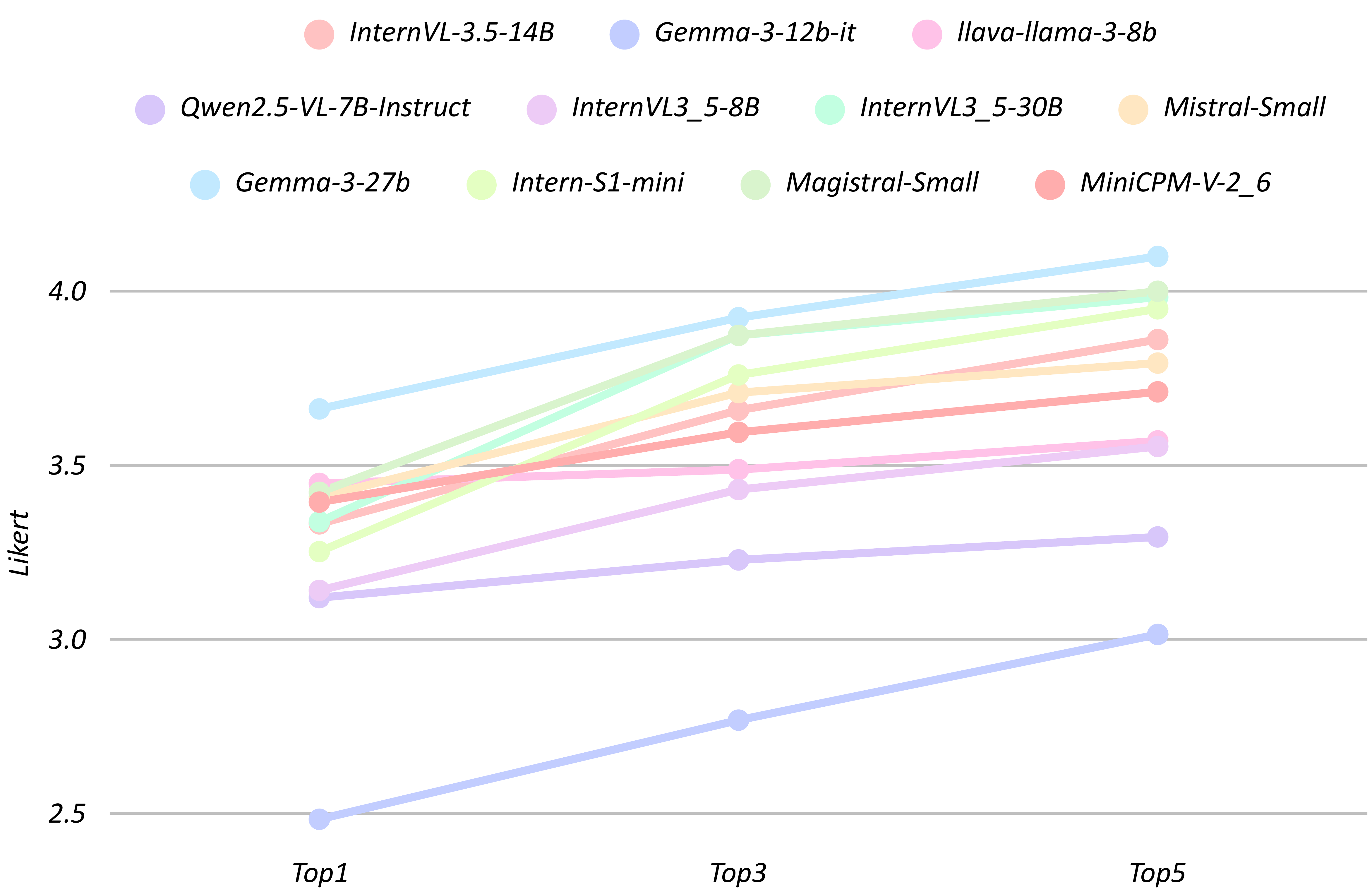}
    \caption{Retrieval effect of different top-$k$ settings.}
    \label{fig:topk}
\end{figure}

\textbf{Ablation Study.} To assess the contribution of each AVR component, we perform ablation studies on two representative MLLMs: Gemma-3-27B and InternVL3.5-30B. As shown in Table~\ref{tab:ablation_avr}, all modules improve fine-grained multimodal comprehension. The {adaptive frame interpolation} strategy consistently boosts both $Recall_{v}$ and $Vis_{use}$ by mitigating information loss through dynamic sampling. Further integrating {OCR} and {DET} tools yields additional gains, with OCR particularly effective for structured scenes involving textual or numeric content.

\textbf{Performance on Multi-hop Questions.}
To further investigate model performance in complex scenarios that require integrating fine-grained visual cues for multi-step reasoning, we take Qwen2.5-VL-7B-Instruct as an base model and compare the 1–5 Likert~\cite{joshi2015likert} score distribution for multi-hop questions with (Multi-hop*) or without AVR (Multi-hop). Higher Likert scores indicate better balanced use of multimodal information and more complete coverage of keypoints. As shown in Fig.~\ref{fig:hop}, incorporating AVR leads to a noticeable increase in average Likert scores, particularly for score 4, demonstrating that AVR enables the model to better comprehend fine-grained content, which in turn supports more accurate multimodal multi-hop reasoning.

\textbf{Impact of Retrieval Top-$k$.} As shown in Fig.~\ref{fig:topk}, we also conducted experiments comparing different retrieval top-$k$ settings, using the Likert score as the evaluation metric. The Likert~\cite{joshi2015likert} score provides a holistic assessment of answer clarity, balanced use of multimodal information, and coverage of keypoints. We observe that as $k$ increases from 1 to 5, the performance of nearly all models improves, indicating that retrieving more relevant segments is generally beneficial for the fine-grained video-based MRAG task. However, the magnitude of improvement varies across models, suggesting that different MLLMs have varying abilities to capture and utilize fine-grained multimodal information.

\begin{figure}[t]
    \centering
    \includegraphics[width=0.9\linewidth]{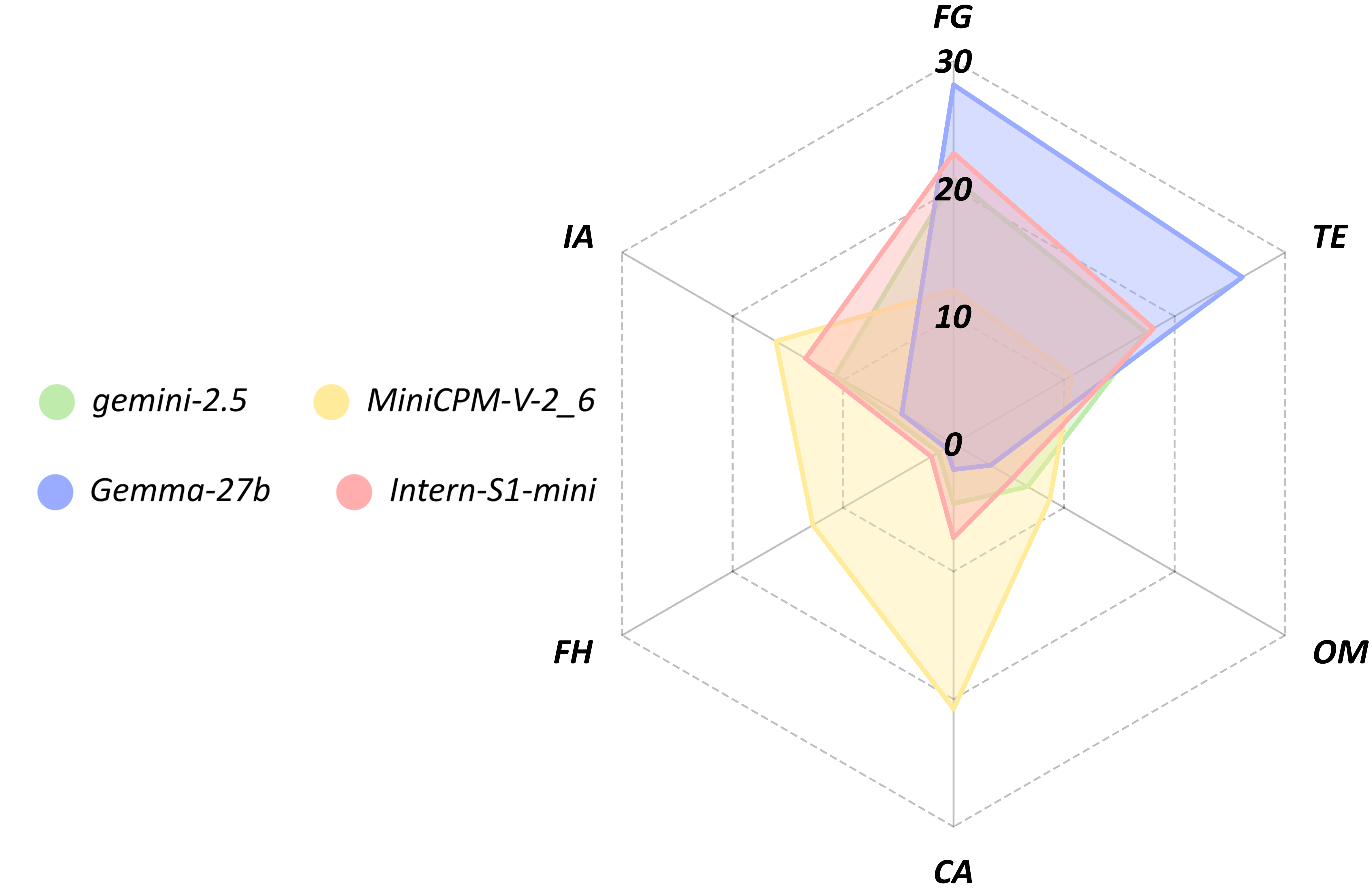}
    \caption{Human evaluation on CFVBench with AVR.}
    \label{fig:human_eval_with_AVR}
\end{figure}

\textbf{Human Evaluation.} Following the same protocol as Section~\ref{sec:eval}, we further evaluated 100 responses from 4 representative MLLMs after integrating the AVR framework. Each response was independently annotated by 3 human raters, achieving a Fleiss’s kappa~\cite{fleiss1973equivalence} of 0.83, indicating strong agreement. Majority voting results are shown in Fig.~\ref{fig:human_eval_with_AVR}, we find: (1) compared to methods without AVR in Fig.~\ref{fig:human_value_raw}, both FG and TE errors of all MLLMs notably decreased, validating AVR’s effectiveness in capturing fine-grained visual cues via adaptive frame selection and OCR/DET tools; while (2) hallucination and contextual association errors persist, particularly for MiniCPM-V-2\_6, suggesting that intrinsic reasoning instability remains even when visual details are accurately captured. A case study experiment was also conducted, please refer to Appendix~\ref{appendix:case}.


\section{Conclusions}
In this work, we introduce CFVBench, a large-scale, human-checked benchmark of 599 real-world videos and 5,360 open-ended QA pairs spanning tables, charts, and other formats. The benchmark covers audio-text-video modalities and is designed to evaluate fine-grained multimodal comprehension in existing video-based MRAG systems.  Comprehensive evaluation of 7 retrieval methods and 14 MLLMs reveals that current approaches struggle with transient but critical multimodal details. To address this, we propose Adaptive Visual Refinement (AVR), a simple yet effective framework that combines adaptive frame sampling with on-demand tool invocation, consistently enhancing fine-grained understanding and overall performance across all the evaluated MLLMs. CFVBench and AVR could provide a practical testbed and methodology for advancing video-based multimodal reasoning in real-world scenarios.
\bibliographystyle{ACM-Reference-Format}
\bibliography{sample-base,refs}

\appendix

{\section{The Full-sized Example in CFVBench}
\label{appendix:full_size}
The full-sized example in Fig.~\ref{fig:intro} from the Introduction section is presented in Fig.~\ref{fig:full_size}.

\begin{figure*}
    \centering
    \includegraphics[width=0.9\linewidth]{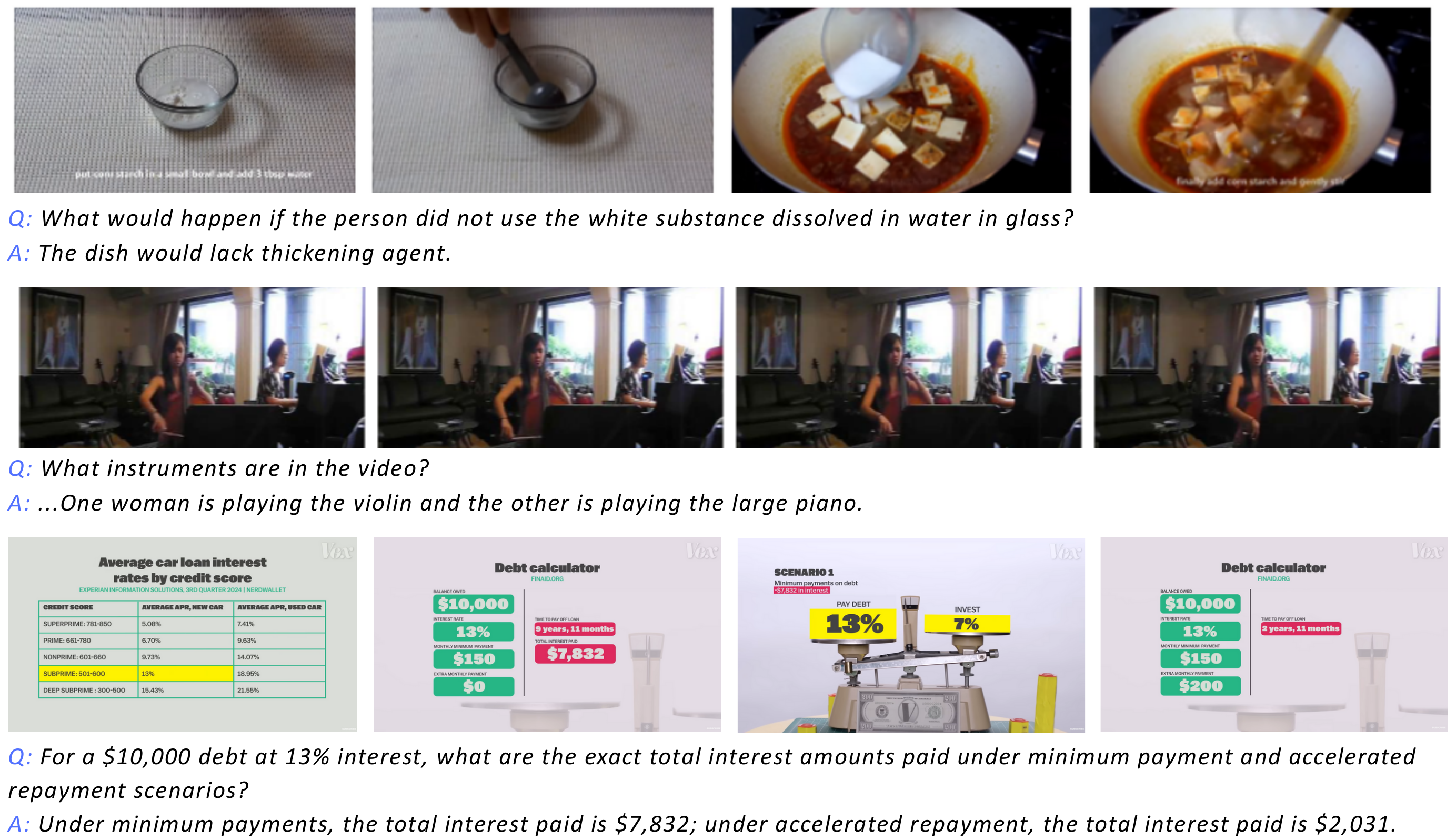}
    \caption{A full-sized cases comparison between existing video-based MRAG benchmarks and CFVBench. In CFVBench, crucial clues are embedded in tables and images of video frames, requiring MLLMs to perform fine-grained reasoning.}
    \label{fig:full_size}
\end{figure*}

\section{CFVBench Data Source}
\label{appendix:source}
The data source of different topics of CFVBench is in Table~\ref{tab:dataset-source}.

\begin{table}
  \caption{Data Source of CFVBench.}
  \label{tab:dataset-source}
  \resizebox{0.4 \textwidth}{!}{  
  \begin{tabular}{cc}
    \toprule
   Topic & Website Link \\
    \midrule
    Structured-Data Videos & \href{https://www.youtube.com/@Vox}{Vox} \\
    Tutorials & 
    \begin{tabular}[t]{@{}c@{}}
      \href{https://www.youtube.com/playlist?list=PL_dhPga7ruufIAJuSAXu_9O47Q2WfQ4xQ}{Adobe Character Animator} \\
      \href{https://www.youtube.com/playlist?list=PL_dhPga7ruueJ4GWDtBGf67G6S-QWp8wp}{LosslessCut} \\
      \href{https://www.youtube.com/playlist?list=PL_dhPga7ruufetjMfOoWhWAN2LrscdEP9}{Wix ADI}
    \end{tabular} \\
    News & \href{https://www.youtube.com/@Reuters}{Reuters} \\
    \bottomrule
  \end{tabular}
  }
\end{table}

\section{Statistics of CFVBench}
\label{appendix: Statistics}
CFVBench is a large-scale, manually curated multimodal benchmark for evaluating fine-grained reasoning in video-based MRAG systems. It contains 5,360 open-ended QA pairs from 599 real-world YouTube videos across diverse domains such as news, reports, and tutorials, with an average duration of 232.3 seconds. Questions are concise (15.22 words on average) and include 3,703 single-hop and 1,660 multi-hop items, the latter averaging 2.72 reasoning keypoints. “What” and “How” questions dominate (48.13\% and 25.62\%), while other types such as “Which,” “Why,” and “Where” contribute additional diversity.


\begin{table}
  \caption{CFVBench properties and question  statistics.}
  \label{tab:dataset-stats}
  \small
  \begin{tabular}{cc}
    \toprule
     Metric & Value \\
    
    \midrule
    Average Video Length & 232.3s \\
    Average Question Length & 15.22 words \\
    Total Questions & 5,360 \\
    Single-hop Questions & 3,703 \\
    Multi-hop Questions & 1,660 \\
    Average Keypoints per Multi-hop Question & 2.72 \\
    \midrule
    What & 2,581 (48.13\%) \\
    How & 1,374 (25.62\%) \\
    Mixed & 719 (13.41\%) \\
    Which & 251 (4.68\%) \\
    Why & 239 (4.46\%) \\
    Where & 68 (1.27\%) \\
    When & 59 (1.10\%) \\
    Who & 30 (0.56\%) \\
    Other & 42 (0.78\%) \\
    \bottomrule
  \end{tabular}
\end{table}

\section{Robustness Experiment}
\label{appendix:Robustness}
We further evaluate AVR’s robustness via a reorder experiment, in which the five retrieved video clips are randomly shuffled. As shown in Table~\ref{tab:rerank}, models with AVR consistently outperform their base counterparts across F1, St\_cos, and Likert scores. For example, Gemma-3-27B shows a 6.26\% F1 and 2.68\% St\_cos improvement, indicating that AVR reliably enhances fine-grained reasoning even when the input clip order is disrupted.


\section{Case Study}\label{appendix:case}
\begin{figure*}[t!]
    \centering
    \includegraphics[width=0.8\linewidth]{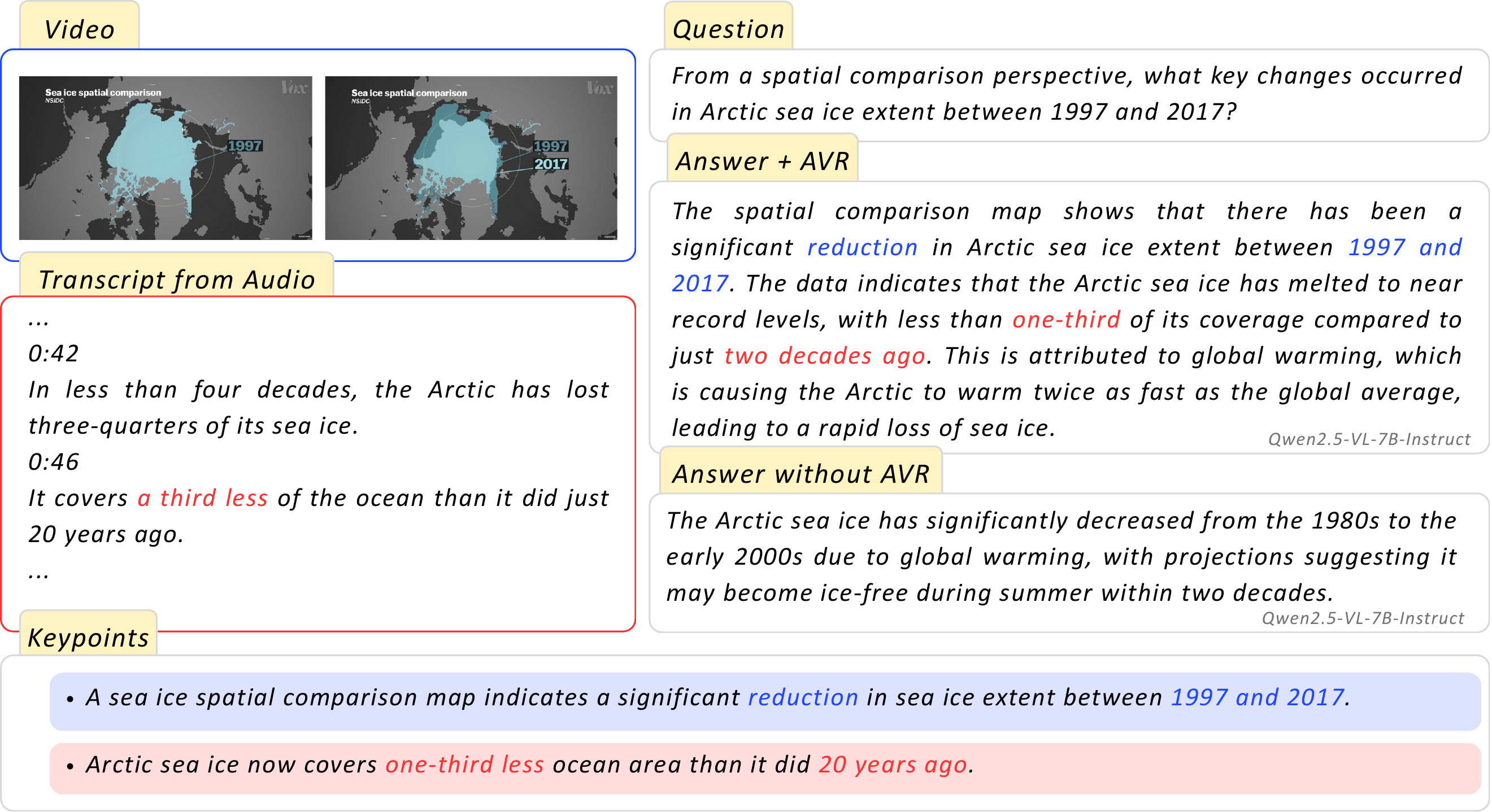}
    \caption{The case study experiment, where key clues are embedded in the on-screen text and audio transcript of the video.}
    \label{fig:casestudy}
\end{figure*}

To qualitatively illustrate AVR’s capabilities, we present a case study with Qwen2.5-VL-7B-Instruct (Figure~\ref{fig:casestudy}), where key clues appear in on-screen text and the audio transcript. Without AVR, the baseline fails to link multimodal evidence and produces an incorrect answer, demonstrating hallucination. With AVR, the model successfully integrates multimodal cues to generate a complete, accurate, and coherent answer, showing that AVR effectively resolves incomplete evidence and enhances fine-grained visual comprehension.


\begin{table}
  \caption{Effect of reordering on generation quality. \textcolor{softgreen}{Green} values indicate relative changes compared to the base models.}
  \label{tab:rerank}
  \centering
  \resizebox{0.49 \textwidth}{!}{  
  \begin{tabular}{c c c c}
    \toprule
    Model & F1 & St$\_$cos & Likert \\
    \midrule
    Gemma-3-12b-it  & 0.0748 \textcolor{softgreen}{+1.63\%} & 0.6128 \textcolor{softgreen}{+1.37\%} & 2.9580 \textcolor{softgreen}{+4.88\%} \\
    Qwen2.5-VL-7B-Instruct  & 0.1181 \textcolor{softgreen}{+5.54\%} & 0.6295 \textcolor{softgreen}{+1.16\%} & 3.2078 \textcolor{softgreen}{+3.49\%} \\
    InternVL3$\_$5-14B  & 0.1688 \textcolor{softgreen}{+2.18\%} & 0.6824 \textcolor{softgreen}{+0.15\%} & 3.5294 \textcolor{softgreen}{+1.03\%} \\
    llava-llama-3-8b  & 0.1368 \textcolor{softgreen}{+3.32\%} & 0.7969 \textcolor{softgreen}{+0.45\%} & 3.5630 \textcolor{softgreen}{+1.52\%} \\
    Mistral-Small  & 0.1393 \textcolor{softgreen}{+1.24\%} & 0.7059 \textcolor{softgreen}{+0.91\%} & 3.7529 \textcolor{softgreen}{+1.15\%} \\
    InternVL-3.5-8B  & 0.1756 \textcolor{softgreen}{+2.87\%} & 0.7307 \textcolor{softgreen}{+0.48\%} & 3.7850 \textcolor{softgreen}{+2.10\%} \\
    Intern-S1-mini  & 0.2108 \textcolor{softgreen}{+3.89\%} & 0.7882 \textcolor{softgreen}{+3.07\%} & 3.9059 \textcolor{softgreen}{+1.93\%} \\
    InternVL3$\_$5-30B  & 0.1515 \textcolor{softgreen}{+0.53\%} & 0.7882 \textcolor{softgreen}{+5.88\%} & 3.9412 \textcolor{softgreen}{+2.91\%} \\
    Gemma-3-27b  & 0.1681 \textcolor{softgreen}{+6.26\%} & 0.8235 \textcolor{softgreen}{+2.68\%} & 4.0235 \textcolor{softgreen}{+1.34\%} \\
    MiniCPM-V-2$\_$6  & 0.2165 \textcolor{softgreen}{+13.47\%} & 0.7882 \textcolor{softgreen}{+1.49\%} & 3.6824 \textcolor{softgreen}{+1.21\%} \\
    \bottomrule
  \end{tabular}
  }
\end{table}

\section{Prompts}\label{appendix:prompt}
The dataset, code, and prompts will be released upon acceptance of the paper.

\end{document}